\documentclass[final]{cvpr}

\usepackage{times}
\usepackage{epsfig}
\usepackage{graphicx}
\usepackage{amsmath}
\usepackage{amssymb}
\usepackage{ragged2e}
\usepackage{multirow}

\def\ie{\textit{i.e.}~}
\def\etal{\textit{et al.}~}

\makeatletter
\def\blfootnote{\xdef\@thefnmark{}\@footnotetext}
\makeatother

\usepackage[pagebackref=true,breaklinks=true,colorlinks,bookmarks=false]{hyperref}

\def\cvprPaperID{6128} %
\def\confYear{CVPR 2021}

\begin{document}

\title{PSRR-MaxpoolNMS: Pyramid Shifted MaxpoolNMS with Relationship Recovery }
\author{
{Tianyi Zhang$^{1}$ \qquad Jie Lin$^{1*}$ \qquad Peng Hu$^{2}$ \qquad Bin Zhao$^{3}$ \qquad Mohamed M. Sabry Aly$^{4}$}\\

$^{1}$ I2R, A*star, Singapore  $^{2}$ Sichuan University, China $^{3}$ IME, A*star, Singapore 
$^{4}$ NTU, Singapore\\

{ $\{$zhang\_tianyi, lin-j$\}$@i2r.a-star.edu.sg,  penghu.ml@gmail.com,
 zhaobin@ime.a-star.edu.sg, msabry@ntu.edu.sg}

}

\maketitle
\pagestyle{empty}
\thispagestyle{empty}

\begin{abstract}
Non-maximum Suppression (NMS) is an essential post-processing step in modern convolutional neural networks for object detection. Unlike convolutions which are inherently parallel, the de-facto standard for NMS, namely GreedyNMS, cannot be easily parallelized and thus could be the performance bottleneck in convolutional object detection pipelines. MaxpoolNMS is introduced as a parallelizable alternative to GreedyNMS, which in turn enables faster speed than GreedyNMS at comparable accuracy. However, MaxpoolNMS is only capable of replacing the GreedyNMS at the first stage of two-stage detectors like Faster-RCNN. There is a significant drop in accuracy when applying MaxpoolNMS at the final detection stage, due to the fact that MaxpoolNMS fails to approximate GreedyNMS precisely in terms of bounding box selection. In this paper, we propose a general, parallelizable and configurable approach PSRR-MaxpoolNMS, to completely replace GreedyNMS at all stages in all detectors. By introducing a simple Relationship Recovery module and a Pyramid Shifted MaxpoolNMS module, our PSRR-MaxpoolNMS is able to approximate GreedyNMS more precisely than MaxpoolNMS. Comprehensive experiments  show that our approach outperforms MaxpoolNMS by a large margin, and it is proven faster than GreedyNMS with comparable accuracy. For the first time, PSRR-MaxpoolNMS provides a fully parallelizable solution for customized hardware design, which can be reused for accelerating NMS everywhere.
\end{abstract}

\blfootnote{*Corresponding author: J. Lin (lin-j@i2r.a-star.edu.sg)\\
This research is supported by the Agency for Science, Technology and Research (A*STAR) under its AME Programmatic Funds (Project No.A1892b0026).}

\section{Introduction}

\begin{figure}[!t]
\begin{center}
  \includegraphics[width=0.95\linewidth]{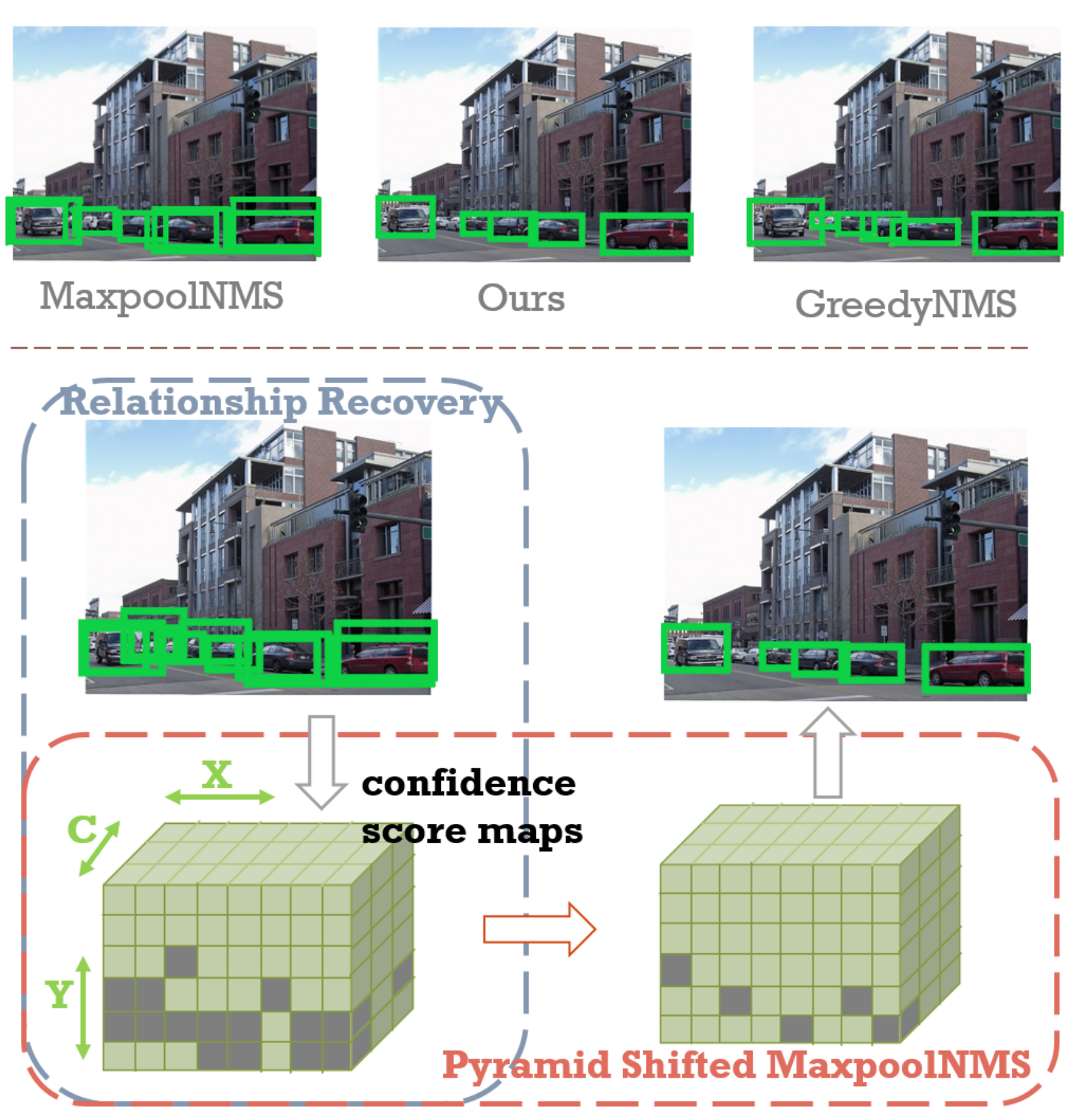}
\end{center}
  \caption{(Top) Visualized comparison of MaxpoolNMS~\cite{cai2019maxpoolnms}, our method and GreedyNMS at the final detection stage of Faster-RCNN. Compared to MaxpoolNMS, our method behaves more like GreedyNMS.
  (Bottom) Pipeline of PSRR-MaxpoolNMS. Relationship Recovery to build up the confidence score maps, followed by Pyramid Shifted MaxpoolNMS to eliminate overlapped boxes and only keep the boxes with peak scores. Each cell on the map encodes the confidence score, scale/ratio ($C$) and spatial location ($X,Y$) of bounding boxes.
  }
\label{fig:brief_pipeline}
\vspace{-1mm}
\end{figure}

Object detection is one of the key tasks in computer vision, with the objective of localizing and classifying objects in a scene. During the past few years, deep convolutional neural networks has emerged as the champion of object detection~\cite{girshick2015fast,ren2015faster,liu2016ssd}. Convolutional object detectors are broadly grouped into either one-stage detectors like SSD~\cite{liu2016ssd} and YOLO~\cite{redmon2016you} or two-stage detectors like Faster RCNN~\cite{ren2015faster} and R-FCN~ \cite{dai2016r}, in which convolutions often account for the majority of computing operations. On the other side, significant progress has been made towards better-performing dedicated hardware for accelerating the convolution operations by exploiting their inherent parallelism, such as GPU and Google TPU~\cite{jouppi2017datacenter}. Therefore, the execution time spent on convolution operations is decreasing rapidly, e.g., at milliseconds.

Non-maximum Suppression (NMS), as a must-have post-processing technique in all convolutional object detectors, is likely to become the performance bottleneck in object detection pipelines~\cite{cai2019maxpoolnms}. The de-facto standard for NMS, namely GreedyNMS, is composed of a sorting operation over confidence scores for tens of thousands of bounding boxes, followed by nested for loops to greedily select the boxes with high scores and remove the boxes significantly overlapped with the selected ones. Unlike convolutions which are inherently parallel, GreedyNMS cannot be easily parallelized due to the nested for loops. Thus, GreedNMS would gradually dominate the execution time of convolutional object detectors~\cite{cai2019maxpoolnms}, as convolutions run faster thanks to the increasing parallelism on dedicated hardware (e.g., from P100 to V100 GPU).

MaxpoolNMS~\cite{cai2019maxpoolnms}, as a parallelizable alternative to GreedyNMS, is introduced to largely accelerate GreedyNMS without incurring loss in detection accuracy.
MaxpoolNMS is inspired by the observation that bounding boxes with high confidence scores correlate to peak values on the so-called confidence score maps in which the spatial relationship between anchor boxes is preserved. Therefore, NMS can be designed as simple max pooling on the score maps which encode confidence scores, scales, ratios, and spatial locations of anchor boxes (see Fig~\ref{fig:brief_pipeline} Bottom). After max pooling, only boxes with peak scores are kept and the others are suppressed. In terms of execution time, MaxpoolNMS runs much faster than GreedyNMS, mainly attributed to the fact that max pooling operations are inherently parallel.

However, MaxpoolNMS is dedicated only to replacing the GreedyNMS at the first stage of two-stage detectors, e.g., the GreedyNMS after the region proposal network in Faster-RCNN. There is a significant drop in detection accuracy when directly applying MaxpoolNMS at the second stage of two-stage detectors, e.g., the GreedyNMS after the detection network in Faster-RCNN (see visualized example in Fig.~\ref{fig:brief_pipeline} Top and quantitative results in Table~\ref{table:maxpoolNMS}).  This lowers the value of MaxpoolNMS since a customized hardware for MaxpoolNMS cannot be reused to replace GreedyNMS at all stages in all detectors.

1

In this paper, we propose a general approach, namely PSRR-MaxpoolNMS, to completely replace GreedyNMS at all stages in all detectors. The key is to approximate GreedyNMS as precisely as possible, which can be measured by the overlap ratio of the selected bounding boxes between GreedyNMS and the approximation method. As evidenced by the low overlap ratio (see Table~\ref{table:maxpoolNMS}), MaxpoolNMS fails to approximate GreedyNMS, mainly due to (A) the score map mismatch problem on confidence score maps (see Fig.~\ref{fig:mismatch}) and (B) the difficulty of maximizing the score map sparsity (see Fig.~\ref{fig:pyramidMaxpoolNMS} and  Fig.~\ref{fig:shiftedMaxpoolNMS}) with a single-scan max pooling on the confidence score maps. PSRR-MaxpoolNMS introduces a Relationship Recovery module and a Pyramid Shifted MaxpoolNMS module to solve problem (A) and (B) respectively. As a result, our PSRR-MaxpoolNMS is able to approximate GreedyNMS more precisely than MaxpoolNMS (see the overlap ratio in Table~\ref{table:maxpoolNMS}).

We summarize our contributions as follows: 
\begin{itemize}
\item A general approach PSRR-MaxpoolNMS to accelerate NMS at all stages in all convolutional object detectors.

\item A Relationship Recovery module to correct the score map mismatch when projecting bounding boxes to confidence score maps, enabling more accurate scale, aspect ratio and spatial relationships between boxes. 

\item A Pyramid Shifted MaxpoolNMS on confidence score maps to significantly increase the sparsity of the score maps, and thus eliminate more overlapped boxes.

\item In PSRR-MaxpoolNMS, the Relationship Recovery and the Pyramid Shifted MaxpoolNMS are simple and parallelizable operations. Therefore, for the first time, PSRR-MaxpoolNMS provides a fully parallelizable solution for customized hardware design, which can be reused for accelerating NMS everywhere.

\item Finally, our PSRR-MaxpoolNMS outperforms MaxpoolNMS by a large margin. Moreover, it is proven faster than GreedyNMS with comparable accuracy.

\end{itemize}

\section{Related Works}

\subsection{One-stage and Two-stage Object Detectors}

Convolutional object detection frameworks are roughly classified into One-stage detectors and Two-stage detectors. One-stage detectors like SSD and YOLO~\cite{liu2016ssd,redmon2016you,lin2017focal} directly predict the bounding box coordinates and class probability by passing an entire image through a single unified network.
Two-stage detectors like Faster-RCNN~\cite{ren2015faster,he2017mask,girshick2015fast} are based on the class agnostic region proposals. The region proposals are the  candidate bounding boxes that potentially enclose target objects.  Different from previous works RCNN \cite{girshick2014rich} or Fast-RCNN \cite{girshick2015fast} which employ hand-crafted region proposal generation \cite{uijlings2013selective}, Faster-RCNN \cite{ren2015faster} generates region proposals by training a Region Proposal Network (RPN). 
The features of the proposals are fed into subsequent detection network to predict the final box coordinates and class-specific probability for each proposal.

\subsection{Non-maximum Suppression}
The final goal of object detectors is to output exactly one bounding box to tightly enclose each target object. However, most object detection pipelines tend to generate redundant highly-overlapped bounding boxes to enclose an object, hence introducing large number of false positives. Non-maximum Suppression (NMS) is an essential step to suppress the redundant bounding boxes.
The most widely used NMS method is GreedyNMS \cite{dalal2005histograms}. GreedyNMS firstly sorts the boxes by their confidence scores in descending order, then iteratively selects the most confident predictions from the remaining boxes and eliminates all the other boxes that have large overlap with the selected ones.
There are variants of NMS to increase the detection accuracy \cite{bodla2017soft,liu2019adaptive,hosang2016convnet, gahlert2020visibility,salscheider2020featurenms}. SoftNMS~\cite{bodla2017soft} decreases the scores of the boxes to be suppressed, instead of deleting these boxes by hard thresholding. Adaptive NMS~\cite{liu2019adaptive} learns to adaptively set the box selection threshold according to object density. Hosang \etal~\cite{hosang2016convnet} reformulates NMS as ConvNet that can be trained end to end. Visibility guided NMS~\cite{gahlert2020visibility} leverages the detection of the whole objects as well as the detection of the visible parts to tackle the problem of highly occluded object detection. FeatureNMS~\cite{salscheider2020featurenms} leverages on the feature embedding distance to determine whether to suppress or keep the candidate boxes.

Hardware-aware NMS acceleration has been less explored. MaxpoolNMS~\cite{cai2019maxpoolnms} reformulates NMS as max pooling on confidence score maps to remove the redundant boxes. Max pooling operations are inherently parallel, thus MaxpoolNMS is much more efficient than GreedyNMS which cannot be easily parallelized. However, MaxpoolNMS is only confined to the region proposal network (RPN) of the two-stage detectors, and cannot be generalized to all stages in all detectors including one-stage detectors.

\section{Method}

In this section, we first briefly review MaxpoolNMS \cite{cai2019maxpoolnms} (Section~\ref{sec:revisit}) and analyze its limitations (Section~\ref{sec:limitation}). Then we introduce our PSRR-MaxpoolNMS to address the limitations. PSRR-MaxpoolNMS is composed of two steps: Relationship Recovery (Section~\ref{sec:recovery}), followed by Pyramid Shifted MaxpoolNMS (Section~\ref{sec:pyramidshifted}).

\subsection{Revisiting MaxpoolNMS}
\label{sec:revisit}

\begin{figure}[t]
\begin{center}
  \includegraphics[width=0.8 \linewidth]{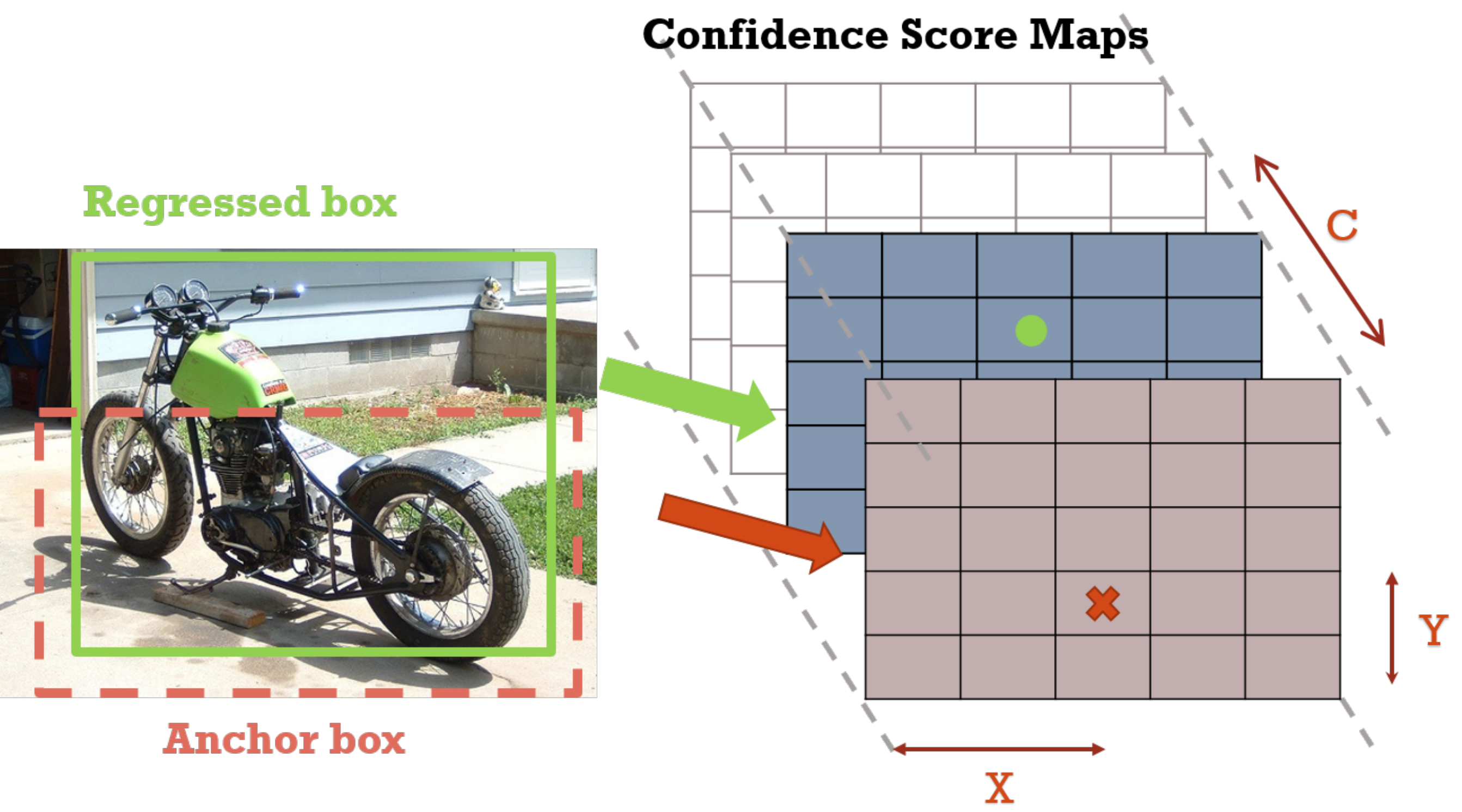}
\end{center}
  \caption{MaxpoolNMS~\cite{cai2019maxpoolnms} projects a anchor box (Dashed red) to a score map with ratio 1:2 (Coral), without consideration of box regression. This leads to the \textbf{score map mismatch} problem, i.e. the regressed box (Solid green) of the anchor has changed in ratio, and thus is projected to another score map with ratio 1:1 (Blue gray). The projection of the regressed box is correct as it encloses the motorcycle more accurately than its corresponding anchor box.}

\label{fig:mismatch}
\end{figure}

MaxpoolNMS~\cite{cai2019maxpoolnms} is an effective yet efficient NMS approach which is specifically designed for removing the overlapped anchor boxes at the first stage of Faster-RCNN detection pipeline, i.e., the Region-Proposal Network (RPN). MaxpoolNMS is composed of two modules. First, as illustrated in Fig.~\ref{fig:mismatch}, it constructs a set of confidence score maps, of which each score map corresponds to a specific combination of anchor box scale and ratio (i.e., channel $c$), and each cell on the score map encodes the objectness score (i.e., cell value) and spatial location (i.e. $x$ and $y$ on the map) of an anchor box that generated by the RPN. For instance, if we use 4 anchor box scales $\{64^2, 128^2, 256^2, 512^2\}$ and 3 anchor box ratios $\{1:2, 1:1, 2:1\}$ for a RPN with down sampling ratio $\beta$ (e.g., $\beta=16$), there are 12 confidence score maps with width 
$\lfloor\frac{W}{\beta}\rceil$ and height $\lfloor\frac{H}{\beta}\rceil$, where $W$ and $H$ denote image width and height respectively.

Second, based on the observation that objects correspond to peak scores on the confidence score maps, a simple max pooling is operated on the maps to suppress anchor boxes with low scores and only keep anchor boxes with peak scores. Moreover, since each score map is dedicated to a specific anchor box size, the kernel size and pool stride for max pooling on that map are determined by its associated anchor box size,
\begin{equation}
\label{eq:kernal_x}
 k_x,s_x = \max(\lfloor\frac{\alpha w}{\beta}\rceil,1),\quad
 k_y,s_y = \max(\lfloor\frac{\alpha h}{\beta}\rceil,1)
\end{equation}
where $k_x$,$k_y$ are the kernal sizes and $s_x$,$s_y$ are the pool strides in $x$ direction and $y$ direction respectively. $w$,$h$ denote the anchor box size (width and height) on a specific score map. $\beta$ is the down sampling ratio for the score maps. $\alpha$ represents the overlap threshold, which is used to control the trade-off between precision and recall. A larger $\alpha$ would suppress more overlapped boxes (lead to higher precision), but at the risk of missed object detections (lead to lower recall). 

Moreover, the max pooling in MaxpoolNMS has 3 variants:
(1) \textbf{Single-Channel MaxpoolNMS}, or multi-scale MaxpoolNMS, applies max pooling on each score map (channel) independently. 
(2) \textbf{Cross-Ratio MaxpoolNMS} concatenates score maps at different ratios for each scale, followed by 3D max pooling on the concatenated maps.
(3) \textbf{Cross-Scale MaxpoolNMS} concatenates score maps at adjacent scales for each ratio, followed by 3D max pooling on the concatenated maps.

Finally, the anchor boxes remaining on the score maps are combined and sorted by their scores in descending order. Only the top boxes are returned as final detections.

\subsection{Limitations of MaxpoolNMS}
\label{sec:limitation}

Though the execution of MaxpoolNMS is much faster than GreedyNMS by paralleling the simple max pooling operations, MaxpoolNMS suffers from a huge shortcoming that it is dedicated only to replacing GreedyNMS at the first stage of regular two-stage convolutional object detectors. To maintain high detection accuracy, GreedyNMS is still a must-have post processing method for the second stage of two-stage detectors and one-stage detectors such as SSD. This makes MaxpoolNMS less attractive in the sense that it lowers the value of a customized hardware for MaxpoolNMS which cannot be easily reused for accelerating NMS at all stages in all detectors.

We observe the detection accuracy drops significantly when applying MaxpoolNMS at the second stage of two-stage detectors. Specifically, we perform MaxpoolNMS after the detection network of Faster-RCNN to remove overlapped boxes, with ResNet-50 as backbone. As shown in Table~\ref{table:maxpoolNMS}, MaxpoolNMS performs significantly worse than GreedyNMS on PASCAL VOC dataset, with over 50\% drop in mAP. As shown in Fig.~\ref{fig:brief_pipeline} Top, one can see that the final selected boxes of MaxpoolNMS is significantly different from that of GreedyNMS, which leads us to a hypothesis that the poor performance of MaxpoolNMS is because it fails to approximate GreedyNMS very well. We measure the quality of the approximation as the overlap ratio of selected bounding boxes between MaxpoolNMS and GreedyNMS. As evidenced in Table~\ref{table:maxpoolNMS}, mAP increases with the overlap ratio, but the overlap ratio for MaxpoolNMS is low.

\begin{figure*}[t]
\begin{center}
  \includegraphics[width=0.85 \linewidth]{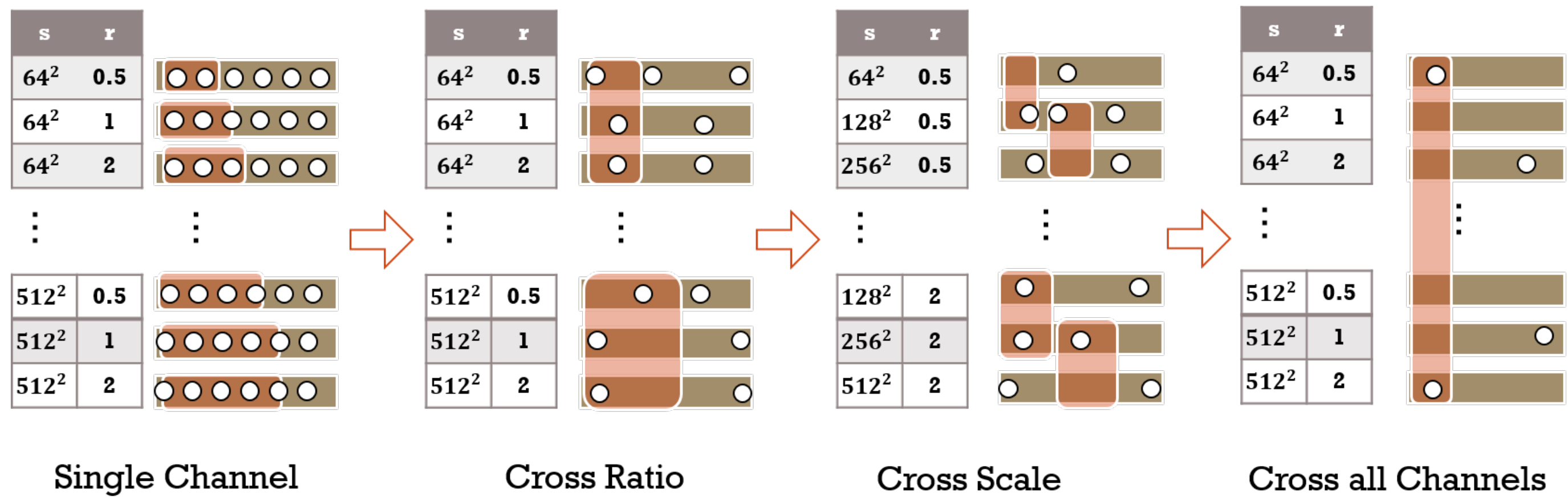}
\end{center}
  \caption{Pyramid MaxpoolNMS. A sequence of max pooling is operated one after another on the confidence score maps, with different channel combinations and pooling parameters (kernel size and stride) determined by the scale ($s$) and ratio ($r$) of the map (From left to right: single channel, cross ratio, cross scale and cross all channels). One can see that the confidence score maps become more and more sparse with the Pyramid MaxpoolNMS (i.e., multi-scan max pooling).}

\label{fig:pyramidMaxpoolNMS}
\end{figure*}

We find that there are 2 key factors that lead to the low overlap ratio for MaxpoolNMS, the score map mismatch on confidence score maps and the difficulty of maximizing the score map sparsity with a single-scan max pooling on the confidence score maps.
\begin{itemize}
\item \textbf{Score map mismatch} occurs during the construction of confidence score maps. MaxpoolNMS projects anchor boxes to score maps without consideration of box regression. This leads to the score map mismatch problem if the regressed boxes correspond to the anchor boxes have changed dramatically in location, scale or aspect ratio. Fig.~\ref{fig:mismatch} shows one example of change in ratio. The mismatch would cause wrong box projections on score maps, which in turn bring in negative effect on the following max pooling operations.

\item \textbf{Low sparsity on score maps}. Since MaxpoolNMS operates only a single-scan max pooling on the confidence score maps, it is hard to achieve high sparsity on dense score maps, implying a lot of highly-overlapped boxes remain after pooling, as illustrated in the left of Fig.~\ref{fig:pyramidMaxpoolNMS} (i.e., max pooling with Single Channel only). Moreover, a single-scan max pooling on the confidence score maps would cause the edge effect. As shown in Fig.~\ref{fig:shiftedMaxpoolNMS} Left, the boxes in the adjacent cells are both kept after max pooling, even though one of them is considered as a duplication.

\end{itemize}

\begin{figure}[b]
\begin{center}
  \includegraphics[width=0.9 \linewidth]{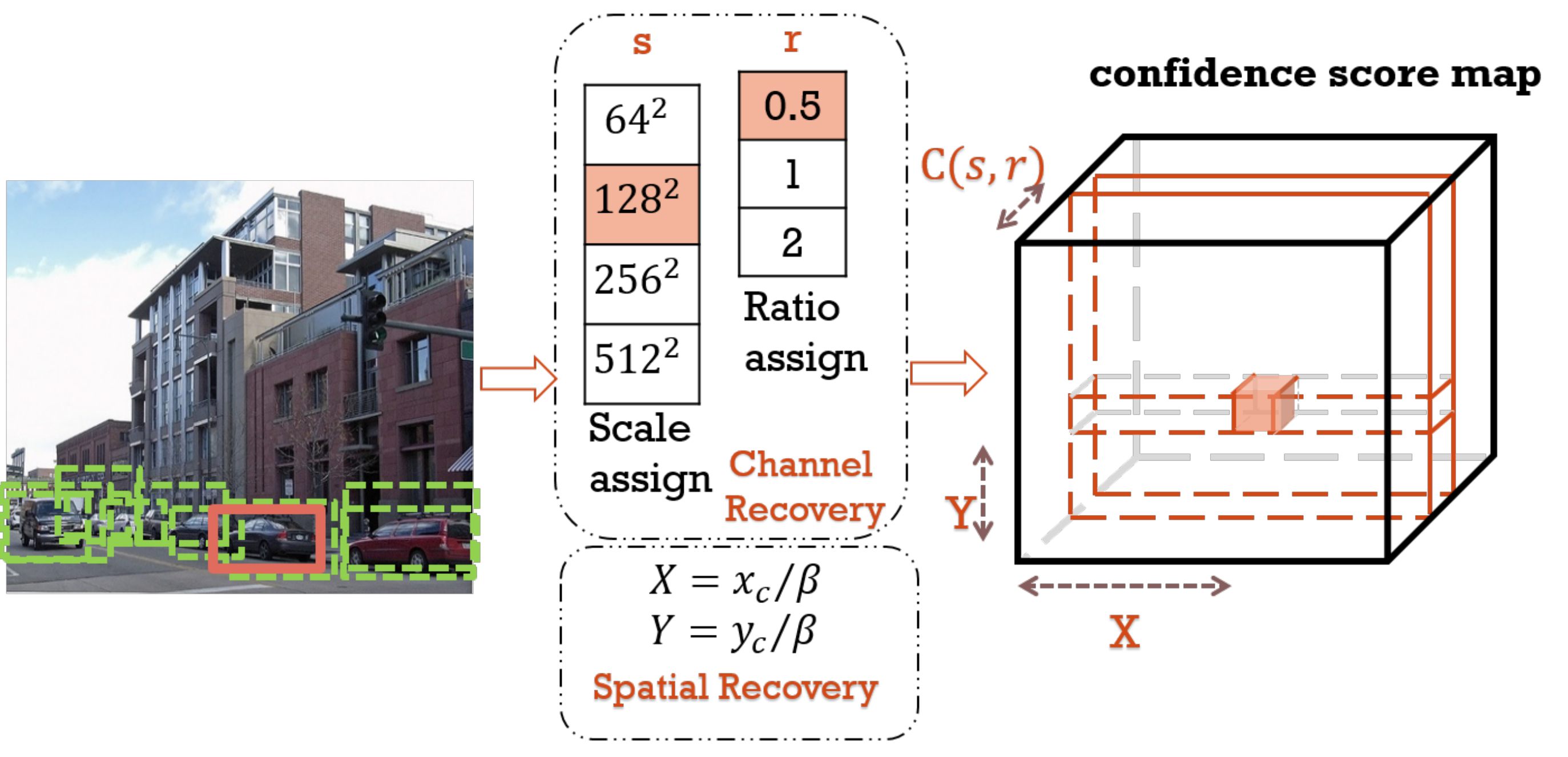}
\end{center}
  \caption{Relationship Recovery to solve score map mismatch.}

\label{fig:indexing}
\end{figure}

\subsection{Relationship Recovery}
\label{sec:recovery}

Instead of projecting anchor boxes to the confidence score maps, our Relationship Recovery projects the regressed boxes to the maps, which solves the score map mismatch problem. With the help of box regression, the regressed boxes in general enclose the objects more accurately than their corresponding anchor boxes, in terms of spatial location, size and shape (i.e., scale and aspect ratio). As such, the confidence score maps projected by the regressed boxes are able to better reflect the actual spatial and channel (a combination of scale and ratio) relationships between objects in a scene. 
Concretely, the Relationship Recovery module consists of three parts: spatial and channel recovery to identify the spatial location ($X,Y$) and the channel ($C(s,r)$) of a regressed box should be mapped to, followed by the score assignment which determines the confidence score for each cell in the maps (see Fig.~\ref{fig:indexing}).

\textbf{Spatial Recovery}.
MaxpoolNMS projects anchor box to wrong spatial location on score map due to dramatic shift of location after box regression. To address this location mismatch problem, given the center position [$x_c,y_c$] of a regressed box in input image, Spatial Recovery maps it to the spatial index [$X,Y$] on score map as
\begin{equation}
\label{eq:spatial_x}
 X = \lfloor \frac{ x_c }{\beta} \rfloor, \quad Y = \lfloor \frac{ y_c }{\beta} \rfloor,
\end{equation}
where $\beta$ is the down sampling ratio of the score maps.

\textbf{Channel Recovery}.
MaxpoolNMS projects anchor box to a channel ($C(s_0,r_0)$) of the score maps simply based on the default scale ($s_0$) and ratio ($r_0$) of the anchor box. Similarly, the channel projection could be wrong if the corresponding regressed box have changed dramatically in scale and/or ratio. To solve this channel mismatch problem, given a regressed box with size $w^{'},h^{'}$, Channel Recovery calculates the nearest scale $s$ to $w^{'} \times h^{'}$ and the nearest ratio $r$ to $\frac{h^{'}}{w^{'}}$ based on Euclidean-distance, and choose $C(s,r)$ as the projected channel for the box.

\textbf{Score Assignment}.
After the spatial locations and channels for all boxes are determined, each cell in the maps could have more than one box projected to it. Therefore, score assignment is introduced to only keep the box with the highest score in each cell. One may note that the score assignment is basically $1\times1$ max pooling in each cell of the score maps, thus it can be treated as a pre-filtering step for removing overlapped boxes that are easy to be identified.

\textbf{Remarks}. All operations of the relationship recovery method are simple and highly parallelizable. In addition, the relationship recovery method is anchor-free. In other words, relationship recovery, as the first step of our PSRR-MaxpoolNMS, opens up a possibility to extend PSRR-MaxpoolNMS from anchor-based one-stage or two-stage convolutional object detectors to anchor-free convolutional object detectors ~\cite{law2018cornernet,duan2019centernet}, since the construction of confidence score maps doesn't reply on anchor box at all. Instead, it only requires the location and size of regressed box, which is accessible in anchor-free detectors as well. We would leave it as our future work.

\subsection{Pyramid Shifted MaxpoolNMS}
\label{sec:pyramidshifted}

We propose Pyramid Shifted MaxpoolNMS to remove overlapped boxes on the confidence score maps, in which the Pyramid MaxpoolNMS aims to thoroughly suppress overlapped boxes across channels (scale and ratio), while the Shifted MaxpoolNMS aims to effectively eliminate overlapped boxes in spatial domain by addressing the edge effect problem. After Pyramid Shifted MaxpoolNMS, the score maps become highly sparse, with only a small number of non-zero cells. The boxes in the non-zero cells are returned as final detections.

\textbf{Pyramid MaxpoolNMS}.
On one hand, MaxpoolNMS operates only a single-scan max pooling on the confidence score maps. On the other hand, MaxpoolNMS assumes overlapped boxes only exist in the channels with adjacent scales (or ratios) on the score maps, which is not always true as the overlapped boxes can be distributed at arbitrary scales/ratios (e.g. a mini cooper occluded by a truck). As such, a single-scan max pooling with invalid assumption is not sufficient to suppress overlapped boxes effectively, resulting in low sparsity on the score maps after pooling. 
One can increase the overlap threshold $\alpha$ in Eq.~\ref{eq:kernal_x} to induce higher sparsity, at the risk of missed detections~\cite{cai2019maxpoolnms}.

We propose Pyramid MaxpoolNMS to induce score map sparsity progressively by executing a sequence of max pooling one after another on the confidence score maps with different channel combinations, as illustrated in~Fig.~\ref{fig:pyramidMaxpoolNMS}. The sequence of max pooling starts from a Single-Channel max pooling, followed by Cross-Ratio and Cross-Scale max pooling, and ends at Cross-all-Channels max pooling. As introduced in Section~\ref{sec:revisit}, Single-Channel max pooling operates on single score map independently, while Cross-Ratio and Cross-Scale max pooling operate on multiple score maps by concatenating channels at adjacent ratios/scales. In addition, we introduce Cross-all-Channels max pooling which operates pooling on all channels. In this way, our Pyramid MaxpoolNMS gradually increases the "receptive field" of pooling operator from local (single score map) to global (all maps), thus without the need of any assumption on the distribution of overlapped boxes.

When operating max pooling on single channel independently, the kernel size and stride for each channel are set as Eq.~\ref{eq:kernal_x}.
When operating max pooling across multiple channels, the kernel size (or stride) is set as the minimum of kernel sizes (or strides) of the channels concatenated. First, if the kernel size is larger than the minimum value, it may suppress true positives detected by the precedent Single-Channel max pooling. Second, the larger gap between scales/ratios, the less likely to have overlapped boxes, hence a small kernel size (or stride) could reduce the risk of suppressing true positives.

\begin{figure}[t]
\begin{center}
  \includegraphics[width=0.8 \linewidth]{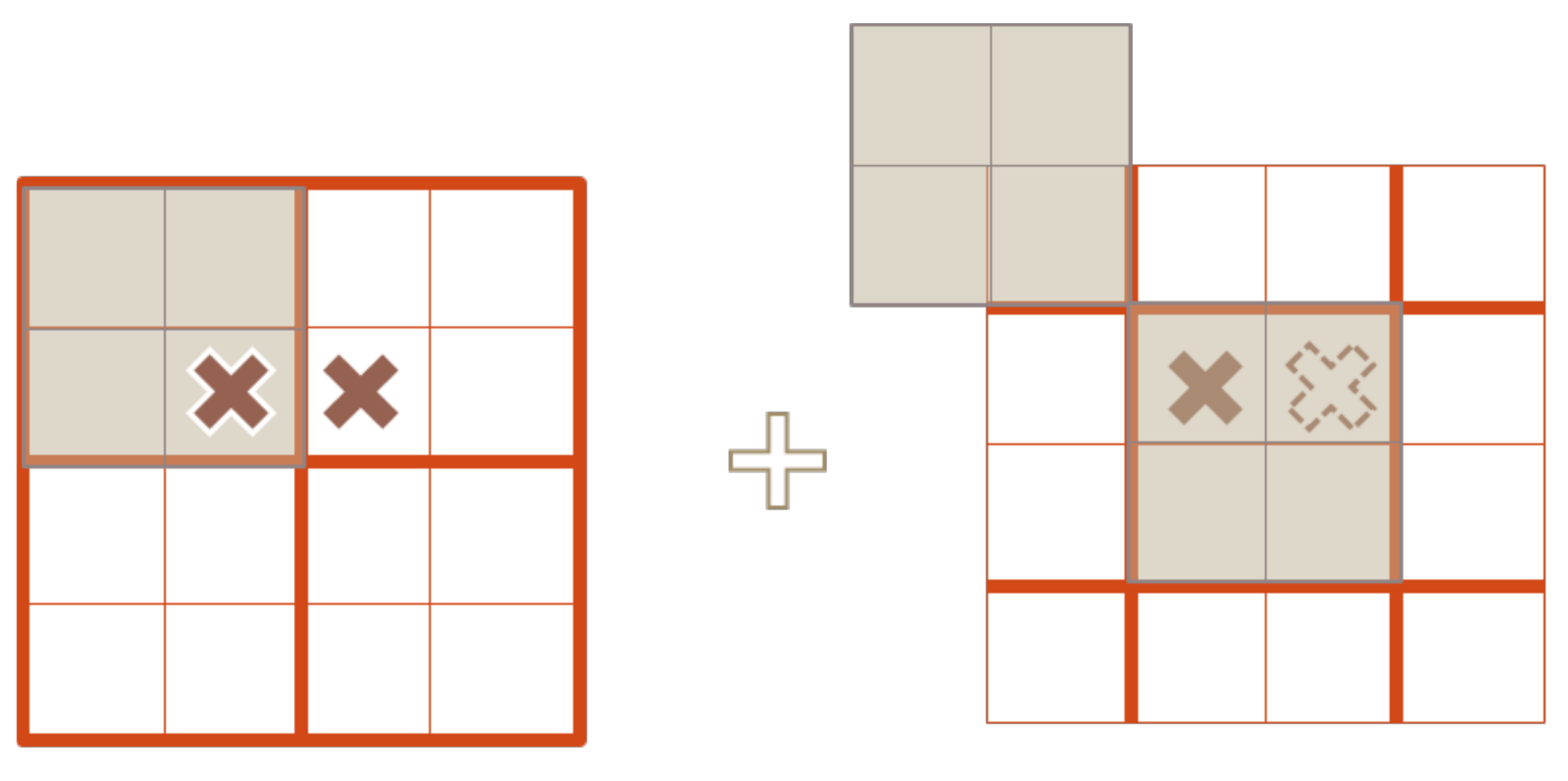}
\end{center}
  \caption{Shifted MaxpoolNMS to alleviate edge effect. (Left) The boxes in the adjacent cells are both kept after max pooling with pool size 2x2 and stride 2. (Right) By adding another max pooling with 1 cell shift, the box with higher score is kept, while the other one is suppressed.}

\label{fig:shiftedMaxpoolNMS}
\end{figure}

\textbf{Shifted MaxpoolNMS}.
Shifted MaxpoolNMS can further increase the score map sparsity, and thus eliminate overlapped boxes in spatial domain ($X,Y$) more effectively.
This is achieved by introducing additional max pooling with a spatial shift on the confidence score maps, which in turn addresses the edge effect problem, as shown in Fig.~\ref{fig:shiftedMaxpoolNMS}. Specifically, given a kernel size $k$, the shifted max pooling is operated on the score maps padded with $\lfloor\frac{ k }{2}\rceil$ zeros around the border. Finally, the shifted max pooling can be appended after each pooling step in the sequence of Pyramid MaxpoolNMS.

\section{Experiments}
\label{sec:exp}

\subsection{Experimental Setup}
\label{subsec:setup}

We evaluate different NMS approaches only at the inference stages of both Faster-RCNN \cite{ren2015faster} and SSD \cite{liu2016ssd}. 
(1) Faster-RCNN \cite{ren2015faster} is a two-stage convolutional object detector. We use ResNet-50, ResNet-101 and ResNet-152~\cite{he2016deep} as the backbone network architectures. For Faster-RCNN training, we follow the default training parameters of the public PyTorch implementation \footnote[1]{https://github.com/jwyang/faster-rcnn.pytorch}. 
Since MaxpoolNMS~\cite{cai2019maxpoolnms} can be viewed as a simplified version of our PSRR-MaxpoolNMS, we simply replace GreedyNMS with Multi-Scale (or Single-Channel) MaxpoolNMS as the post processing technique at the first stage of Faster-RCNN, which achieves comparable accuracy but runs much faster than GreedyNMS.
(2) SSD \cite{liu2016ssd} is a one-stage convolutional object detector. We use VGG-16~\cite{simonyan2014very}, MobileNet-v1~\cite{howard2017mobilenets}, MobileNet-v2~\cite{sandler2018mobilenetv2} as the backbone. We evaluate NMS using the pre-trained models provided by the public PyTorch implementation \footnote[2]{https://github.com/qfgaohao/pytorch-ssd}.  In pre-processing stage, SSD first filters out the bounding boxes with score $<$0.01 for each class. 
For GreedyNMS, it further selects 200 boxes with top scores from the boxes passed the pre-processing stage. Our PSRR-MaxpoolNMS takes all boxes passed the pre-processing stage as input.

For both Faster-RCNN and SSD, our PSRR-MaxpoolNMS is applied to suppress the final bounding box predictions. We fix $\alpha$ to the value of 0.75. In our channel-recovery step, we set the anchors to the scales of  $ \left[ 64 ^2, 128^2, 256^2, 512^2 \right]$ and the ratios of $ \left[ 0.5,1,2 \right]$. 
For Cross-Ratio MaxpoolNMS, all the 3 ratios of each scale are concatenated for max pooling. For Cross-Scale MaxpoolNMS, we only concatenate 2 channels with adjacent scales for each step of max pooling. We denote Single-Channel MaxpoolNMS,  Cross-Ratio MaxpoolNMS and Cross-Scale MaxpoolNMS as \textit{single}, \textit{ratio} and \textit{scale}, respectively.

We perform experiments on PASCAL VOC~\cite{everingham2010pascal} and KITTI~\cite{geiger2012we} datasets. For PASCAL VOC, we train Faster-RCNN detection model using 2007 and 2012 trainval datasets and evaluate on 2007 test dataset. We report the  mean average precision (mAP) for PASCAL VOC dataset.
For KITTI, we randomly split the dataset into 5611 training images and 1870 testing images. We report mAP at difficulty levels from easy to difficult on KITTI. 

\subsection{Comparison with MaxpoolNMS}
\label{subsec:vsmaxpoolnms}

We compare MaxpoolNMS \cite{cai2019maxpoolnms} with our PSRR-MaxpoolNMS approach at the second stage of Faster-RCNN. Results on PASCAL VOC dataset are shown in Table~\ref{table:maxpoolNMS}.
First, we observe that MaxpoolNMS performs poorly, e.g. MaxpoolNMS single 33\% versus GreedyNMS 78.1\%. As expected, though the detection mAP increases with the overlap ratio, the overlap ratio of selected boxes between MaxpoolNMS and GreedyNMS is still very low.
Second, our PSRR-MaxpoolNMS better approximates GreedyNMS which is evidenced by the large overlap ratio and comparable detection accuracy with GreedyNMS (less than 1\% drop in mAP). It is worth noting that similar to \cite{cai2019maxpoolnms}, the only parameter to be set for our method is the overlap threshold $\alpha$. Thus, our method PSRR-MaxpoolNMS does not introduce extra parameter tuning workload, while outperforms MaxpoolNMS by a significant margin.

\begin{table}[t]
\begin{center}
\caption{Detection accuracy (mAP) of our method and MaxpoolNMS on Pascal VOC, at the second stage of Faster-RCNN with ResNet-50 as backbone. We also report the overlap ratio of selected bounding boxes between GreedyNMS and MaxpoolNMS (or our method). As a reference, mAP for GreedyNMS is 78.1\%.}
\label{table:maxpoolNMS}
{
\begin{tabular}{|l|c|c|}
\hline
Method & Box Overlap ($\%$) & mAP ($\%$)\\
\hline
MaxpoolNMS \cite{cai2019maxpoolnms} single  & 15.0 & 33.0\\
MaxpoolNMS \cite{cai2019maxpoolnms} ratio & 18.5 & 36.6 \\
MaxpoolNMS \cite{cai2019maxpoolnms} scale & 11.6 & 26.5 \\
\hline
Ours & 45.3 & 77.6 \\
\hline

\end{tabular}
}
\end{center}
\vspace{-20pt}

\end{table}

\begin{table*}[t]

\begin{center}
\caption{Comparisons of our method and GreedyNMS on KITTI dataset, with ResNet variants as the backbone of Faster-RCNN.}
\label{table:kitti}
{
\begin{tabular}{|c|c|c|c|c|}

\hline

 & & Car & Pedestrian & Cyclist\\
Method & mAP(easy to hard) & Easy \quad Mod \quad Hard  & Easy \quad Mod \quad Hard & Easy \quad Mode \quad Hard\\
 \hline
GreedyNMS & 94.7\quad 89.8\quad 83.0 & {99.1} \quad 97.0 \quad87.4 & {90.2} \quad {81.5} \quad {74.8}  & {94.8} \quad  90.5 \quad  86.9 \\

  Ours (ResNet-50)  & 93.4\quad 88.5\quad 82.8 & 96.4 \quad  95.6 \quad  87.9 & 90.1 \quad  80.9 \quad  74.7 &  93.6 \quad  89.0 \quad  85.7 \\

 \hline

  GreedyNMS & 94.0\quad 88.6\quad 81.5 & {98.6} \quad  95.9 \quad  86.4 & {89.7} \quad  {81.4} \quad  {74.1}  & {93.7} \quad  88.7 \quad  84.1 \\
  Ours (ResNet-101)  & 93.5\quad 88.1\quad 81.2 & 95.9 \quad  95.5 \quad  86.1 & 89.5 \quad  79.5 \quad  72.1 &  95.1 \quad  89.1 \quad  85.2 \\
 \hline
    GreedyNMS & 94.6\quad 89.8\quad 83.0 & {98.3} \quad  95.7 \quad  86.2 & {91.0} \quad  {83.2} \quad  {76.7}  & {94.4} \quad  90.5 \quad  86.1 \\
  Ours (ResNet-152)  & 93.8\quad 89.5\quad 82.7 & 96.8 \quad  96.1 \quad  86.9 & 90.7 \quad  82.8 \quad  75.6 &  93.8 \quad  89.5 \quad  85.6 \\
 \hline

\end{tabular}
}
\end{center}
\end{table*}

\subsection{Comparisons with GreedyNMS}
\label{subsec:vsgreedynms}

We compare our PSRR-MaxpoolNMS with GreedyNMS, on various datasets and convolutional object detectors.
First, we perform experiments on KITTI dataset with Faster-RCNN detector and report detection results in Table~\ref{table:kitti}. 
We observe that our method achieves comparable detection accuracy with GreedyNMS on KITTI at most of the operating points, regardless of the backbone models used.   
Second, we perform experiments on PASCAL VOC dataset with both two-stage (\ie, Faster-RCNN) and one-stage (\ie, SSD) detectors.
As shown in Table~\ref{table:ssd}, 
one can see that our approach performs slightly worse (less than 1\% for most of the operating points) than GreedNMS with various backbone models and different object detectors. 
With Faster-RCNN detector and ResNet-152 as the backbone, the performance gap in mAP between PSRR-MaxpoolNMS and GreedyNMS is only 0.3$\%$. 
It is also worth noting that our PSRR-MaxpoolNMS approach is applicable to various object detection pipelines.

\begin{table}[t]
\begin{center}
\caption{Comparisons of our method and GreedyNMS on PASCAL VOC dataset, with both two-stage detector Faster-RCNN and one-stage detector SSD.}
\label{table:ssd}
{
\begin{tabular}{|l|c|c|}
\hline
 Detection Pipeline & GreedyNMS & Ours \\
 \hline
   Faster-RCNN \cite{ren2015faster} (ResNet-50)& 78.1 & 77.6\\
  Faster-RCNN \cite{ren2015faster} (ResNet-101)& 78.4 & 78.0\\
  Faster-RCNN \cite{ren2015faster} (ResNet-152)& 78.7 & 78.4\\

\hline
   SSD \cite{liu2016ssd} (VGG-16) & 77.3 & 76.1\\
   SSD \cite{liu2016ssd} (MobileNet-v1) & 67.6 & 66.4\\
   SSD \cite{liu2016ssd} (MobileNet-v2) & 68.7 & 67.8\\

 \hline

\end{tabular}
}
\end{center}
\end{table}

\begin{figure}[t]
\begin{center}
  \includegraphics[width=0.9 \linewidth]{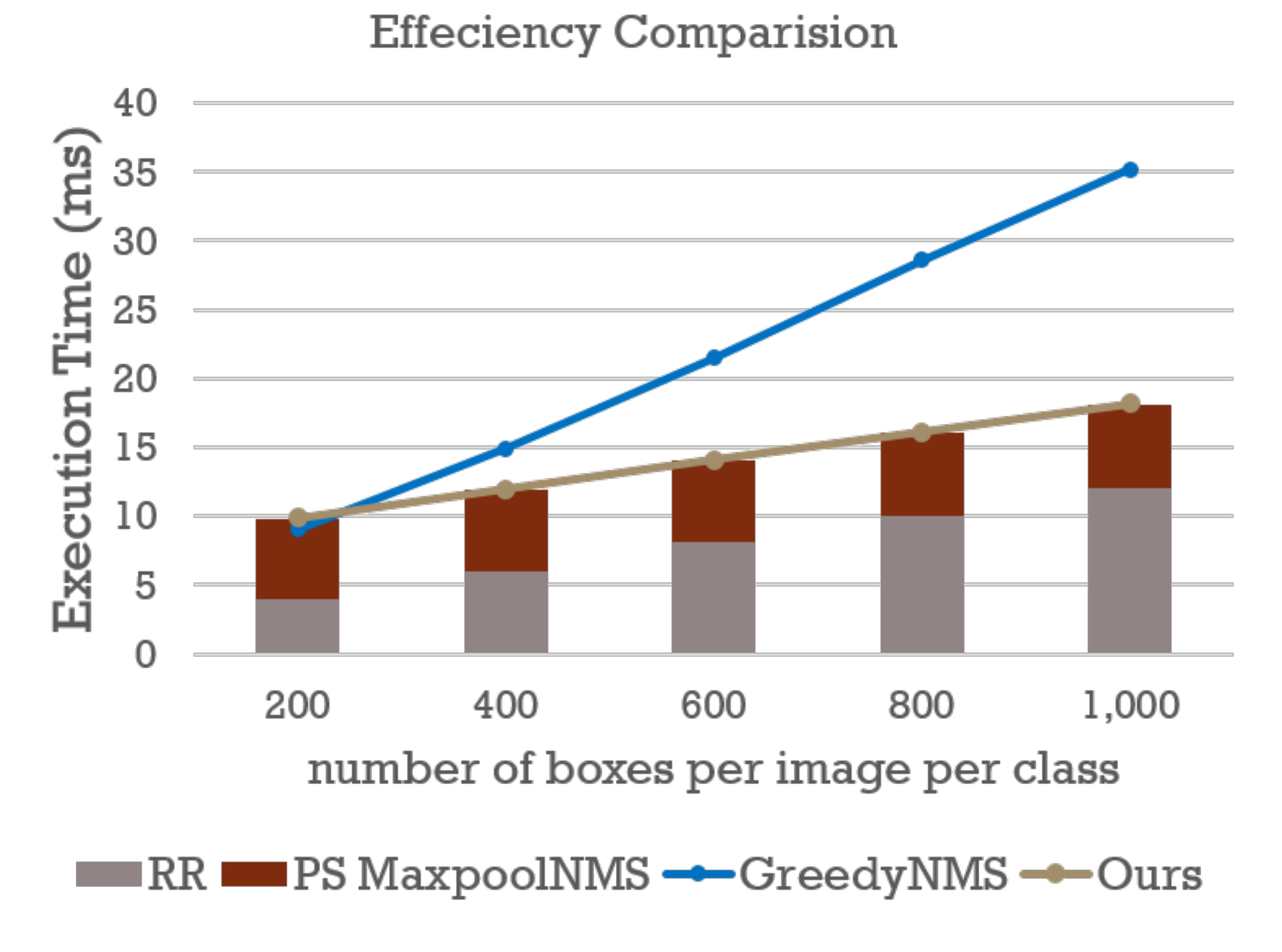}
\end{center}
  \caption{Execution time (in ms) of our method and GreedyNMS as a function of the number of bounding boxes being processed. We also report the timing breakdown of our method (Relationship Recovery (RR), and Pyramid Shifted (PS) MaxpoolNMS). Both methods run on CPU.}

\label{fig:Efficency}
\vspace{-10pt}
\end{figure}

\subsection{Efficiency}
\label{subsec:time}

We perform both theoretical and experimental analysis on the computing efficiency of our method PSRR-MaxpoolNMS. 
Table~\ref{table:timecomplexity} provides the theoretical analysis of the time complexity for GreedyNMS and our method. Given the number of input boxes $N$, the time complexity for both the Relationship Recovery and Pyramid Shifted MaxpoolNMS is $\mathcal{O}(N)$, which is much smaller than that of the GreedyNMS $\mathcal{O}(N\log N) + \mathcal{O}(N^2)$. Moreover, both the Relationship Recovery and Pyramid Shifted MaxpoolNMS can be easily parallelized, which in turn would further reduce the execution time of PSRR-MaxpoolNMS.

We also measure the execution time of GreedyNMS and our method on Intel(R) Core(TM) i9-10900X CPU, with different number of bounding boxes being processed. We experiment on SSD with VGG-16 as backbone on PASCAL VOC dataset. For fair comparison, we remove the score thresholding step in order to set the number of input boxes being processed per image per class. Results are reported in Figure~\ref{fig:Efficency}. First, with the increasing number of input bounding boxes, our PSRR-MaxpoolNMS is more and more efficient than GreedyNMS. When the number of box is increased to 8000, the excutation time of PSRR-MaxpoolNMS is 89 ms while the GreedyNMS takes 512 ms, which is almost 6 times slower than our method. Second, we look into the timing breakdown of PSRR-MaxpoolNMS. We observe that the execution time of Pyramid Shifted MaxpoolNMS is almost constant, while the execution time of Relationship Recovery linearly increases with the number of boxes, due to our own implementation of Relationship Recovery is currently not parallelized. For Pyramid Shifted MaxpoolNMS, we rely on the PyTorch Maxpool and MaxUnpool API which are already parallelized on CPU.

\begin{table}[t]
\begin{center}
\caption{Time Complexity and Parallelism.}
\label{table:timecomplexity}
\resizebox{1.0\linewidth}{!}
{
\begin{tabular}{|c|c|c|}
\hline
 Method & Complexity & Parallel  \\
 \hline
Relationship Recovery & $\mathcal{O}(N)$ & $\checkmark$\\
 PS MaxpoolNMS & $\mathcal{O}(N)$ & $\checkmark$\\
\hline
\hline
GreedyNMS& $\mathcal{O}(N\log N) + \mathcal{O}(N^2)$ & $\times$\\ 
\hline

\end{tabular}
}
\end{center}
\end{table}

\subsection{Ablation studies}
\label{subsec:ablation}

In this section, we perform ablation studies to evaluate different components in our method, i.e., Relationship Recovery and Pyramid  Shifted MaxpoolNMS.
All experimental results are reported on PASCAL VOC dataset, with ResNet-50 as the backbone of Faster-RCNN detector.

\subsubsection{Relationship Recovery}

\textbf{Spatial and Channel Recovery}.
We analyze the effect of each component of Pyramid MaxpoolNMS on spatial and channel recovery.
Results are reported in Table~\ref{table:recovery}. 
One can see that compared with the baseline that is lacking of recovered relationships (because it is based on the anchor box projection), both the spatial and channel recovery step could  alleviate the score map mismatch problem and improve the detection accuracy over the baseline by a large margin, regardless of the channel combination used in the subsequent max pooling stage.

\begin{table}[t]
\begin{center}
\caption{Effect of each component of Pyramid MaxpoolNMS on Spatial and Channel Recovery.}

\label{table:recovery}
{
\begin{tabular}{|c|c|c|c|}
\hline
 - & baseline & spa recover & spa + cha recover\\
 \hline
 single & 41.2 & 44.4 & 71.1\\
  ratio & 49.0 & 63.0 & 73.6 \\
 scale & 31.3 & 67.5 & 68.0 \\
 all & 38.9 & 63.8 & 63.9 \\

 \hline

\end{tabular}
}
\end{center}
\end{table}

\textbf{Score Assignment}.
As mentioned before, for score assignment in each cell of the score maps, we only keep the box with
the highest score, which can be treated as a pre-filtering step based on max pooling in each cell (max-assign).
We also investigate alternatives beyond max-assign, i.e., random-assgin and sum-assign.
'random-assign' is to randomly choose a box (and its score value) that projected to a cell. 'sum-assign' is to sum the score values of all boxes projected to a cell. Like max-assign, both random-assign and sum-assign could be easily parallelized.
Table~\ref{table:duplicate} reports the detection mAP with different score assignment variants. We observe that max-assign performs the best, followed by sum-assign and random-assign.

\begin{table}[t]
\begin{center}
\caption{Effect of the score assignment variants on Relationship Recovery.}
\label{table:duplicate}
{
\begin{tabular}{|l|c|c|c|}
\hline
 method & random-assign & sum-assign &  max-assign\\
 \hline
 mAP ($\%$) & 75.1 & 76.5 & 77.6\\
 \hline

\end{tabular}
}

\end{center}
\end{table}

\subsubsection{Pyramid Shifted MaxpoolNMS}

\textbf{Pyramid MaxpoolNMS}.
We evaluate the effect of the number of channel combinations in a sequence and the execution order in the sequence that defined by Pyramid MaxpoolNMS.
Table~\ref{table:order-channel} reports the detection results.
First, single-scan max pooling (\ie, single, ratio, scale, all.) performs consistently worse than multi-scan max pooling pre-defined in a sequence (e.g., a sequence single+ratio has 2 max pooling), implying the necessity of Pyramid MaxpoolNMS.
Second, if we execute the sequence in reverse order (e.g. for a sequence single+ratio, execute ratio first, followed by single), the performance is slightly worse than the original execution order. 

\begin{table}[t]
\begin{center}
\caption{Effect of the channel combinations and the execution order in the sequence of the Pyramid MaxpoolNMS.}
\label{table:order-channel}
{
\begin{tabular}{|c|c|c|}
\hline
 - & original order & reverse order\\
 \hline
 single (Single-Scale) & 71.1 & -\\
  ratio (Cross-Ratio)  & 73.6 & - \\
 scale (Cross-Scale) & 68.0 & - \\
 all (Cross-All-Channel) & 63.9 & - \\
 \hline

 single+ratio &  74.2 & 73.7\\
 single+ratio+all & 76.9 & 76.1\\
 single+ratio+scale+all & 77.6 & 77.4\\

 \hline

\end{tabular}
}
\end{center}
\end{table}

\begin{table}[t]
\begin{center}
\caption{Effect of the Shifted MaxpoolNMS.}
\label{table:shift}
{
\begin{tabular}{|c|c|c|}
\hline
 - & w/o shift-pool & w/ shift-pool\\
 \hline
  mAP ($\%$) &74.2 & 77.6\\

 \hline

\end{tabular}
}
\vspace{-15pt}

\end{center}
\end{table}

\textbf{Shifted MaxpoolNMS for Edge Effect}.
We perform study on our Shifted MaxpoolNMS and report the results in Table~\ref{table:shift}. It shows that without the additional Shifted MaxpoolNMS to alleviate the edge effect, the detection mAP drops obviously about $3.4\%$.

\section{Conclusion}

In this paper, we propose PSRR-MaxpoolNMS as a parallelizable alternative to GreedyNMS for overlapped box removal in all convolutional object detectors. With the proposed Relationship Recovery module and the Pyramid Shifted MaxpoolNMS module, we tackle the problems of score map mismatch and low sparsity after pooling on the score maps. Comprehensive experiments show that PSRR-MaxpoolNMS achieves comparable detection accuracy with GreedyNMS, but with much higher speedup in execution time.

%
%

%

%

%

%


\begin{thebibliography}{10}\itemsep=-1pt

\bibitem{bodla2017soft}
Navaneeth Bodla, Bharat Singh, Rama Chellappa, and Larry~S Davis.
\newblock Soft-nms--improving object detection with one line of code.
\newblock In {\em Proceedings of the IEEE international conference on computer
  vision}, pages 5561--5569, 2017.

\bibitem{cai2019maxpoolnms}
Lile Cai, Bin Zhao, Zhe Wang, Jie Lin, Chuan~Sheng Foo, Mohamed~Sabry Aly, and
  Vijay Chandrasekhar.
\newblock Maxpoolnms: getting rid of nms bottlenecks in two-stage object
  detectors.
\newblock In {\em Proceedings of the IEEE Conference on Computer Vision and
  Pattern Recognition}, pages 9356--9364, 2019.

\bibitem{dai2016r}
Jifeng Dai, Yi Li, Kaiming He, and Jian Sun.
\newblock R-fcn: Object detection via region-based fully convolutional
  networks.
\newblock In {\em Advances in neural information processing systems}, pages
  379--387, 2016.

\bibitem{dalal2005histograms}
Navneet Dalal and Bill Triggs.
\newblock Histograms of oriented gradients for human detection.
\newblock In {\em 2005 IEEE computer society conference on computer vision and
  pattern recognition (CVPR'05)}, volume~1, pages 886--893. IEEE, 2005.

\bibitem{duan2019centernet}
Kaiwen Duan, Song Bai, Lingxi Xie, Honggang Qi, Qingming Huang, and Qi Tian.
\newblock Centernet: Keypoint triplets for object detection.
\newblock In {\em Proceedings of the IEEE International Conference on Computer
  Vision}, pages 6569--6578, 2019.

\bibitem{everingham2010pascal}
Mark Everingham, Luc Van~Gool, Christopher~KI Williams, John Winn, and Andrew
  Zisserman.
\newblock The pascal visual object classes (voc) challenge.
\newblock {\em International journal of computer vision}, 88(2):303--338, 2010.

\bibitem{gahlert2020visibility}
Nils G{\"a}hlert, Niklas Hanselmann, Uwe Franke, and Joachim Denzler.
\newblock Visibility guided nms: Efficient boosting of amodal object detection
  in crowded traffic scenes.
\newblock {\em arXiv preprint arXiv:2006.08547}, 2020.

\bibitem{geiger2012we}
Andreas Geiger, Philip Lenz, and Raquel Urtasun.
\newblock Are we ready for autonomous driving? the kitti vision benchmark
  suite.
\newblock In {\em 2012 IEEE Conference on Computer Vision and Pattern
  Recognition}, pages 3354--3361. IEEE, 2012.

\bibitem{girshick2015fast}
Ross Girshick.
\newblock Fast r-cnn.
\newblock In {\em Proceedings of the IEEE international conference on computer
  vision}, pages 1440--1448, 2015.

\bibitem{girshick2014rich}
Ross Girshick, Jeff Donahue, Trevor Darrell, and Jitendra Malik.
\newblock Rich feature hierarchies for accurate object detection and semantic
  segmentation.
\newblock In {\em Proceedings of the IEEE conference on computer vision and
  pattern recognition}, pages 580--587, 2014.

\bibitem{he2017mask}
Kaiming He, Georgia Gkioxari, Piotr Doll{\'a}r, and Ross Girshick.
\newblock Mask r-cnn.
\newblock In {\em Proceedings of the IEEE international conference on computer
  vision}, pages 2961--2969, 2017.

\bibitem{he2016deep}
Kaiming He, Xiangyu Zhang, Shaoqing Ren, and Jian Sun.
\newblock Deep residual learning for image recognition.
\newblock In {\em Proceedings of the IEEE conference on computer vision and
  pattern recognition}, pages 770--778, 2016.

\bibitem{hosang2016convnet}
Jan Hosang, Rodrigo Benenson, and Bernt Schiele.
\newblock A convnet for non-maximum suppression.
\newblock In {\em German Conference on Pattern Recognition}, pages 192--204.
  Springer, 2016.

\bibitem{howard2017mobilenets}
Andrew~G Howard, Menglong Zhu, Bo Chen, Dmitry Kalenichenko, Weijun Wang,
  Tobias Weyand, Marco Andreetto, and Hartwig Adam.
\newblock Mobilenets: Efficient convolutional neural networks for mobile vision
  applications.
\newblock {\em arXiv preprint arXiv:1704.04861}, 2017.

\bibitem{jouppi2017datacenter}
Norman~P Jouppi, Cliff Young, Nishant Patil, David Patterson, Gaurav Agrawal,
  Raminder Bajwa, Sarah Bates, Suresh Bhatia, Nan Boden, Al Borchers, et~al.
\newblock In-datacenter performance analysis of a tensor processing unit.
\newblock In {\em Proceedings of the 44th Annual International Symposium on
  Computer Architecture}, pages 1--12, 2017.

\bibitem{law2018cornernet}
Hei Law and Jia Deng.
\newblock Cornernet: Detecting objects as paired keypoints.
\newblock In {\em Proceedings of the European Conference on Computer Vision
  (ECCV)}, pages 734--750, 2018.

\bibitem{lin2017focal}
Tsung-Yi Lin, Priya Goyal, Ross Girshick, Kaiming He, and Piotr Doll{\'a}r.
\newblock Focal loss for dense object detection.
\newblock In {\em Proceedings of the IEEE international conference on computer
  vision}, pages 2980--2988, 2017.

\bibitem{liu2019adaptive}
Songtao Liu, Di Huang, and Yunhong Wang.
\newblock Adaptive nms: Refining pedestrian detection in a crowd.
\newblock In {\em Proceedings of the IEEE Conference on Computer Vision and
  Pattern Recognition}, pages 6459--6468, 2019.

\bibitem{liu2016ssd}
Wei Liu, Dragomir Anguelov, Dumitru Erhan, Christian Szegedy, Scott Reed,
  Cheng-Yang Fu, and Alexander~C Berg.
\newblock Ssd: Single shot multibox detector.
\newblock In {\em European conference on computer vision}, pages 21--37.
  Springer, 2016.

\bibitem{redmon2016you}
Joseph Redmon, Santosh Divvala, Ross Girshick, and Ali Farhadi.
\newblock You only look once: Unified, real-time object detection.
\newblock In {\em Proceedings of the IEEE conference on computer vision and
  pattern recognition}, pages 779--788, 2016.

\bibitem{ren2015faster}
Shaoqing Ren, Kaiming He, Ross Girshick, and Jian Sun.
\newblock Faster r-cnn: Towards real-time object detection with region proposal
  networks.
\newblock In {\em Advances in neural information processing systems}, pages
  91--99, 2015.

\bibitem{salscheider2020featurenms}
Niels~Ole Salscheider.
\newblock Featurenms: Non-maximum suppression by learning feature embeddings.
\newblock {\em arXiv preprint arXiv:2002.07662}, 2020.

\bibitem{sandler2018mobilenetv2}
Mark Sandler, Andrew Howard, Menglong Zhu, Andrey Zhmoginov, and Liang-Chieh
  Chen.
\newblock Mobilenetv2: Inverted residuals and linear bottlenecks.
\newblock In {\em Proceedings of the IEEE conference on computer vision and
  pattern recognition}, pages 4510--4520, 2018.

\bibitem{simonyan2014very}
Karen Simonyan and Andrew Zisserman.
\newblock Very deep convolutional networks for large-scale image recognition.
\newblock {\em arXiv preprint arXiv:1409.1556}, 2014.

\bibitem{uijlings2013selective}
Jasper~RR Uijlings, Koen~EA Van De~Sande, Theo Gevers, and Arnold~WM Smeulders.
\newblock Selective search for object recognition.
\newblock {\em International journal of computer vision}, 104(2):154--171,
  2013.

\end{thebibliography}
{\small
\bibliographystyle{ieee_fullname}

}

\end{document}